\documentclass[conference]{IEEEtran}

\makeatletter
\def\ps@IEEEtitlepagestyle{%
 \def\@oddfoot{\mycopyrightnotice}%
 \def\@evenfoot{}%
}

\makeatletter
\def\ps@IEEEtitlepagestyle{%
  \def\@oddfoot{\mycopyrightnotice}%
  \def\@oddhead{\hbox{}\@IEEEheaderstyle\leftmark\hfil\thepage}\relax
  \def\@evenhead{\@IEEEheaderstyle\thepage\hfil\leftmark\hbox{}}\relax
  \def\@evenfoot{}%
}
\def\mycopyrightnotice{%
  \begin{minipage}{\textwidth}
  \centering \scriptsize
  Copyright~\copyright~2022 IEEE. Personal use of this material is permitted. Permission from IEEE must be obtained for all other uses, in any current or future media, including\\reprinting/republishing this material for advertising or promotional purposes, creating new collective works, for resale or redistribution to servers or lists, or reuse of any copyrighted component of this work in other works by sending a request to pubs-permissions@ieee.org.
  \end{minipage}
}
\makeatother

\IEEEoverridecommandlockouts
\usepackage{cite}
\usepackage{amsmath,amssymb,amsfonts}

\usepackage{amsthm}
\usepackage{hyperref}

\usepackage{algorithmic}
\usepackage{graphicx}
\usepackage{textcomp}
\usepackage{xcolor}
\usepackage[flushleft]{threeparttable}
\usepackage[]{algorithm2e}
\usepackage{wrapfig}
\usepackage{capt-of}
\usepackage{caption}
\usepackage{subcaption}
\usepackage{url}

\usepackage{capt-of}  
\usepackage{cuted}    
\newcommand\tab[1][1cm]{\hspace*{#1}}
\usepackage{nopageno}

\begin{document}

\begin{strip}
\begin{huge}
\textbf{IEEE Copyright Notice}\\
\end{huge}

\vspace{5mm} 

\begin{large}
Copyright (c) 2022 IEEE
\end{large}
\vspace{5mm} 
\begin{large}

Personal use of this material is permitted. Permission from IEEE must be obtained for all other uses, in any current or future media, including reprinting/republishing this material for advertising or promotional purposes, creating new collective works, for resale or redistribution to servers or lists, or reuse of any copyrighted component of this work in other works.\\




\textbf{Accepted to be published in: }9th Annual Conference on Computational Science \& Computational Intelligence (CSCI'22: Dec 14-16, 2022, USA)\\
\newline
\newline
Cite as:
\end{large}
\end{strip}
\begin{large}
\begin{tabular}{|l|}
     \hline
     O. Izunna, H. Shane, and K. Jess. “Machine Learning Methods for Evaluating \\ Public Crisis: Meta-Analysis,” 2022 International Conference on Computational \\ Science and Computational Intelligence (CSCI), Las Vegas, NV, USA, 2022. 
     \\
     \hline
\end{tabular}\\
\newline
\newline
\newline
\newline
BibTeX:\\

\begin{tabular}{|l|}
     \hline
        @InProceedings\{okpala2022methods, \\
        \tab author = \{Okpala, Izunna and Halse, Shane and Kropczynski, Jess\}, \\
        \tab title = \{Machine Learning Methods for Evaluating Public Crisis: Meta-Analysis\}, \\
        \tab booktitle = \{2022 International Conference on Computational Science and Computational \\
        \tab Intelligence (CSCI)\}, \\ \tab month = \{December 14 – 16\}, \\
        \tab year = \{2022\}, \\
        \tab publisher = \{IEEE\}, 
        \\
        \} \\
     \hline
\end{tabular}

\end{large}

\vspace{5mm} 

\title{Machine Learning Methods for Evaluating Public Crisis: Meta-Analysis\\
}

\author{\IEEEauthorblockN{1\textsuperscript{st} Izunna Okpala}
\IEEEauthorblockA{\textit{School of Information Technology} \\
\textit{University of Cincinnati}\\
okpalaiu@mail.uc.edu}
\and
\IEEEauthorblockN{2\textsuperscript{nd} Shane Halse}
\IEEEauthorblockA{\textit{School of Information Technology} \\
\textit{University of Cincinnati}\\
halsese@ucmail.uc.edu}
\and
\IEEEauthorblockN{3\textsuperscript{rd} Jess Kropczynski}
\IEEEauthorblockA{\textit{School of Information Technology} \\
\textit{University of Cincinnati}\\
kropczjn@ucmail.uc.edu}


}

\maketitle
\thispagestyle{empty}
\begin{abstract}
This study examines machine learning methods used in crisis management. Analyzing detected patterns from a crisis involves the collection and evaluation of historical or near-real-time datasets through automated means. This paper utilized the meta-review method to analyze scientific literature that utilized machine learning techniques to evaluate human actions during crises. Selected studies were condensed into themes and emerging trends using a systematic literature evaluation of published works accessed from three scholarly databases. Results show that data from social media was prominent in the evaluated articles with 27\% usage, followed by disaster management, health (COVID) and crisis informatics, amongst many other themes. Additionally, the supervised machine learning method, with an application of 69\% across the board, was predominant. The classification technique stood out among other machine learning tasks with 41\% usage. The algorithms that played major roles were the Support Vector Machine, Neural Networks, Naive Bayes, and Random Forest, with 23\%, 16\%, 15\%, and 12\% contributions, respectively.

\end{abstract}


\section{Introduction}
Over the last decade, the scientific community like IEEE and the Information Systems for Crisis Response and Management (ISCRAM) have contributed many studies that utilize real-time information sources to support situation awareness during large-scale events \cite{nguyen2019review,steen2019analysis}. The Machine learning field has advanced on how they automate processes to filter large volumes of data \cite{kabir2021emocov}. This study explores a variety of machine learning solutions utilized in scholarly articles to understand human actions towards crises. It was also informed by studies that addressed disaster management, health, politics, and other forms of crisis that utilized data beyond social media with evidential proof from many scholarly articles available in major academic databases that focused on analyzing human actions. While the first interactive medium for an individual that has no control over the mainstream media is the social media platform
, local sources for tracking crises exist
. People tend to report incidents, or debate about the ongoing incident via a social network that is familiar to them, or verifiable local reporting agencies and news media \cite{hwang2015social,oh2013community}. Scientific researchers have taken advantage of the mass surge of social media data \cite{Nathan2021explore}, and local reports to carry out machine learning procedures like predictions. 
The objective of this paper is to examine prevalent machine learning methods utilized by academics for managing crises, how those methods are implemented, and the source of the data.
 
As noted earlier, historical and real-time data play an important role in managing crisis, or in our case, evaluating crisis \cite{praveen2021analyzing}. In order to plan for, mitigate, and avoid future crises, it is recommended to investigate historical or existing solutions. The concept of human action is predicated on some causative factors i.e. before someone acts, there is a cause \cite{fritts2020actions}. The human environment is the main factor that drives crisis, and how humans manage environmental resources plays a huge role in crisis occurrence. Some of the concepts that aid in determining human actions from a data source are sentiments, perceptions, and/or attitudes \cite{chen2015media}. While perception uniquely identifies opinion-based thoughts and impressions \cite{haupt2021characterizing,okpala2022perception}, and can simply be distinguished from sentiments and attitudes, sentiments emphasizes emotions \cite{sv2021indian,udebuana2019analysis}, and attitudes leads to actions \cite{praveen2021analyzing}. This study seeks to demonstrate with the help of peer-reviewed articles, the machine learning methods that are most prevalent in today’s world for evaluating or predicting crisis. The focus would be on public or human actions towards crisis with key concepts like attitudes and/or perceptions.


\subsection{Research Questions}

\begin{itemize}
\item[\textbf{1.}] \textit{What are the dominant machine learning methods for managing crisis?}

\item[\textbf{2.}] \textit{What are the keywords frequently used in scientific studies addressing crisis?}
\end{itemize}
\section{Background}

Statistical methods were among the first tools used to evaluate crises \cite{young2018model}. In 1932, when Patrick \cite{patrick1932comparison} completed a multivariate analysis on some organizations, financial stakeholders began developing models to assess the likelihood of a crisis in their organization. Since then, academics have devised a number of quantitative ways to detect and evaluate crises. Some quantitative analysis like the t-test has been successful in quantifying ratios \cite{sonmez2007impact}. Altman \cite{altman1968financial} devised a score that was used to categorize observations into good and bad. Multiple Discriminant Analysis (MDA) also played a role in some advanced analyses to compress variance between datasets \cite{ohlson1980financial}. Despite their widespread use in both academia and industry, these types of models have proven to be about numbers and quantities, necessitating the need for improvements that span beyond numbers \cite{song2004comparison}. %
To address the constraint of these models, various research that employs pattern matching approaches has been substantially researched in the field of machine learning \cite{barboza2017machine}. Several of which have proved machine learning models' ability to deal with unbalanced datasets \cite{farquad2012preprocessing}, pictorial data and text data \cite{faure1999knowledge}. Even the difference between parametric and non-parametric methods for analyzing risks can be detected \cite{altman2009parametric}.

In addressing the research questions, the authors explored some background literature related to auto-coding, pattern matching, and text analysis. Most articles tackled issues of crisis management using machine learning, data transformation/scaling, and natural language processing. Machine learning has helped IT practitioners perform tasks in a very short amount of time \cite{chowdhury2013tweet4act}. It appears to be a quick option for identifying disruption events, getting authentic feeds or detecting periodic incidents in real-time \cite{starbird2012learning}. Learning such patterns was also a major game-changer, as the approach tries to map patterns of interest or similarities in a given dataset \cite{szczyrba2020machine}, while also showing the capacity to learn and produce accurate results \cite{angaramo2018online}. %
The pattern that is key to understanding human actions are cues demonstrating preference. This preference can be in the form of emotions, opinions, viewpoints, or specific annotations that help in explaining why people act the way they do \cite{wiggins2003attitudes}. When evaluating or predicting actions, some key concepts to take note of are perceptions, attitudes, sentiments, etc. The term "perception" can be misconstrued to mean the same thing as attitudes or even sentiments. In simpler terms, sentiments are concerned with people's feelings about an event, i.e., positive and negative events \cite{udebuana2019analysis}, perception and attitudes are concerned with people's perspective towards an event \cite{wiggins2003attitudes,okpala2022perception}. Emotions can aid in understanding perception, but the distinction is that perceptions or attitudes can be formed on the basis of facts and not only emotions \cite{landreville2019and}. While attitudes are reactionary (they can produce immediate action), perceptions are internal cues suggesting future actions or attitudes \cite{faronbi2017perception}. Focusing on perceptions, the formal definition is the process of organizing and interpreting sensory inputs to make sense of events \cite{okpala2022perception}. 

To detect or predict a crisis, there is a need to make sure that the dataset used for analysis is actionable. Actionable data is characterized by information that can be acted upon or data that provides sufficient insight about the future \cite{kropczynski2018identifying} i.e., the data gives insight into actions that informs valuable decisions. In other words, it is more than just data kept in data warehouses. They have undergone analytical and data manipulation and are presented in a clear, intelligible, and frequently visually appealing manner \cite{spasojevic2015identifying}. It enables researchers to spot mistakes or potential crises and capitalize on new opportunities, improve future actions, and make faster and more informed decisions for the future \cite{avvenuti2018crismap}.

\subsection{Machine Learning for Crisis Detection and Management}
There are multiple explanations as to why a crisis incident should be analyzed with machine learning. One reason is to prevent such occurrences in the future, another is to study the pattern by which people engage on such occasions in a timely fashion. The COVID-19 epidemic was the major global crisis between 2020 - 2021 \cite{wu2020exposure}, and many academics have attempted to evaluate live or historical data as to how the surge is escalating, including the metrics that supposedly caused the escalation. Other researchers have looked at why some people would desire to get vaccinated \cite{okpala2022perception} and why others would not. Beyond the COVID-19 pandemic, research on crisis informatics tackles all forms of crisis, like disaster management, 911 or 311 incidents, political movements, natural disasters, and various forms of assault like rape, to name a few. Given that our study focuses on crisis situations, machine learning has shown promise in the scientific community, as evident in the number of articles that apply automated means for detecting, predicting, or averting crisis events. Almost all the major unsupervised and supervised algorithms like the neural network, support vector machine, Naïve Bayes algorithm, K-means clustering, K nearest neighbor, decision trees, and gradient boost algorithms have been applied in various capacities. 
Several crisis events can be averted or addressed in a timely manner when trained models with high accuracy are used, especially in cases where human involvement in solving the problem is near impossible or time-consuming. 

Averting dangers has been made easier and more expressive with the help of several data sources and machine learning methods. The Twitter platform is one such medium from which actionable data can be derived. Thousands of academics have explored the platform since it can be used to generate insights during crisis \cite{hellerstein2002discovering}. Additionally, when dealing with data from varied sources, the issue of data structure limits some procedures, but with the proper application of machine learning tuning or data transformation, the issue can be mitigated. Some other data sources however are dedicated to archiving data for certain topics, such as the health crisis - World Health Organization data (WHO) \cite{karo2018world}, financial crisis - World Bank open data \cite{weaver2019open}, imminent disasters in government - data.gov \cite{birchall2015data}, or child mortality and maternal mortality - UNICEF \cite{world2015trends}.

Given a reliable and actionable data source, machine learning thrives at conducting computational tasks that would ordinarily take human intelligence a significant amount of time to handle. Machine learning has evolved over time in such a way that any computer device with memory can be taught to follow specific patterns \cite{stauffer2000learning}. The traditional approach to learning, in which humans are trained with specialized material and tested to determine their mastery of a topic, gave rise to the concept of machine learning. It describes a machine's ability to have some type of intelligence and readiness to learn from experience. The game of checkers is one example of machine learning through experience \cite{samuel1959some}. Beyond the checkers game, machines have been trained to differentiate between authentic news and fake news as well as spam emails from authentic emails using the BERT model \cite{cao2020bilingual}. The value of implementing such machine-ready systems makes the prediction of crisis events seamless.

\section{Methodology}
\begin{figure}
  \centering
  \includegraphics[width=.8\linewidth]{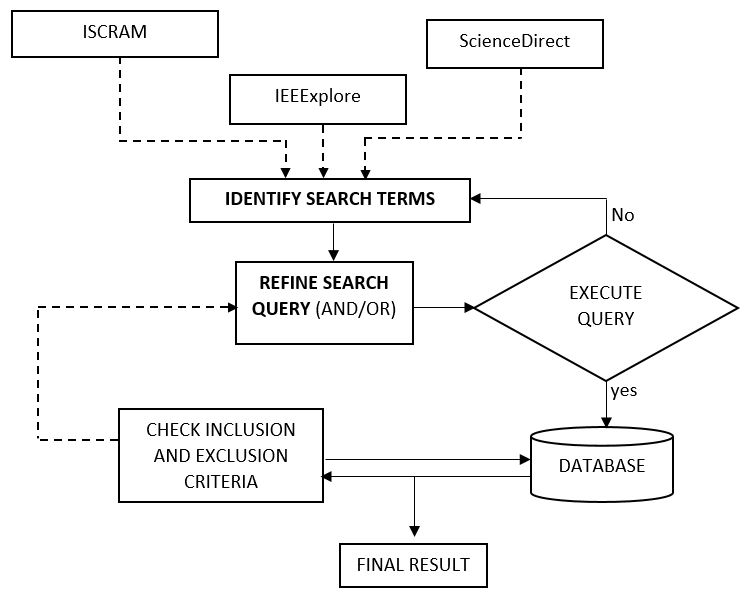}
  \caption{Data search and selection Algorithm}
  \label{fig: s-algo}
\end{figure}

The methodology employed in this study is the meta-review technique. It identifies the feasible variables in a cluster of articles with a common interest. This approach is often applied to literature published in a particular language; in our case, the review focused on academic articles written in English and published in ISCRAM, ScienceDirect, and the IEEEXplore databases. These databases were selected because they have subsections that addressed crisis management using machine learning. That is not to say that some other databases with an emphasis on crisis do not exist; we mainly focused on three for this study. The search and selection criterion illustrated in \autoref{fig: s-algo} shows the various building blocks and the flow of data across them. The search terms are essentially the keywords needed to be used in a query for specific databases. More emphasis on the search term was made in the search technique section. The AND and OR operators were used in the query structure because they help to sieve out the articles that do not fall within our query parameter. \autoref{fig: truncate} expanded the inclusion and exclusion criteria shown in \autoref{fig: s-algo} to visualize the different components of the inclusion and exclusion mechanisms, and how data were truncated or reduced in the process to get a final result.


\subsection{Search technique}

The search technique was critical in this evaluative review to ensure integrity. The paper was found using an automated search in a variety of different electronic databases. \autoref{tab:scope} shows the three scientific databases explored. 

\begin{table}[hbt!]
\begin{center}
    \caption{\label{tab:scope}Scope of the search.}
    \begin{tabular}{|c|c|c|}
         \hline
         Index&NAME&URL\\
         \hline
         \hline
         DB1&IEEEXplore Digital Library&\url{https://ieeexplore-ieee-org/}\\
         \hline
         \hline
         DB2&ScienceDirect Library&\url{https://www.sciencedirect.com/}\\
         \hline
         \hline
         DB3&ISCRAM Digital Library&\url{http://idl.iscram.org/}\\
         \hline
    \end{tabular}
\end{center}
\end{table}

With a structured search pattern, this study aimed at getting only relevant articles targeted to answer our research question i.e., at the search stage, we sieved literature from relevant sources with the selection of appropriate keywords. The articles featured keywords like; machine learning, crisis, and disaster. The steps for keyword preparation are as follows:
\begin{enumerate}
  \item Determine the search terms in relation to the research questions.
  \item Ensure that alternative spellings, antonyms, and synonyms of the search term are identified as well.
  \item Perform Boolean operations (AND, OR) on the search terms.
  \item Identify the dates for the search query.
\end{enumerate}

The following keywords and operators, which are reflective of our research questions appear in the paper: (Disaster) OR (Crisis) AND (Machine Learning).

The search was conducted based on the database's preference pertaining to query structure. The ISCRAM was queried with the Contextual Query Language (CQL) specific to ISCRAM with the code in \autoref{tab:query}. IEEEXplore and Science Direct were less complicated with the help of their web-based portal for inputting the search parameters with the "AND" and "OR" operators as shown in \autoref{tab:query}.


\begin{table}[hbt!]
\begin{center}
    \caption{\label{tab:query}How the Databases were queried}
    \begin{tabular}{|c|c|c|}
         \hline
         Index&Query type&Query\\
         \hline
         \hline
         1&IEEExplore& “Crisis AND Machine learning OR Disaster”\\
         \hline
         \hline
         2&ScienceDirect&“Crisis AND Machine learning OR Disaster”\\
         \hline
         \hline
         3&ISCRAM (CQL)& “all abstract machine learning crisis disaster”\\
         \hline
    \end{tabular}
\end{center}
\end{table}

Additionally, Our search was restricted by publication year, i.e. between 2010 and 2021 (recent publications), as well as categories, which included peer-reviewed journal and conference papers only. 

\subsection{Data Selection}

The 55 articles reviewed were carefully selected using the step-by-step approach shown in \autoref{fig: s-algo}. The years from 2010 to 2021 as stated earlier were used primarily to reflect current trends and progression in machine learning practice with regard to crises. The initial response from the various databases using the query format shown in \autoref{tab:query} were 2,274 articles from Science Direct, 16,236 articles from IEEExplore, and 76 articles from ISCRAM. Given the volume of articles, the number of publications were reduced to emphasize the significance of our research area (crisis). We specifically chose just the journal and conference articles from the computer science field for the Science Direct database because this discipline was clearly prominent with more contributions in the area of machine learning. This resulted in a total of 186 publications. We employed the same strategy for the IEEE search, but the advanced search parameters were different. The initial result of 16,236 from IEEExplore was reduced to 5,770 articles by limiting the topic categories to disasters and choosing only conference and journal articles. When the computer science discipline was applied we got a total of 476 articles.

\begin{figure}[!ht]
  \centering
  \includegraphics[width=.95\linewidth]{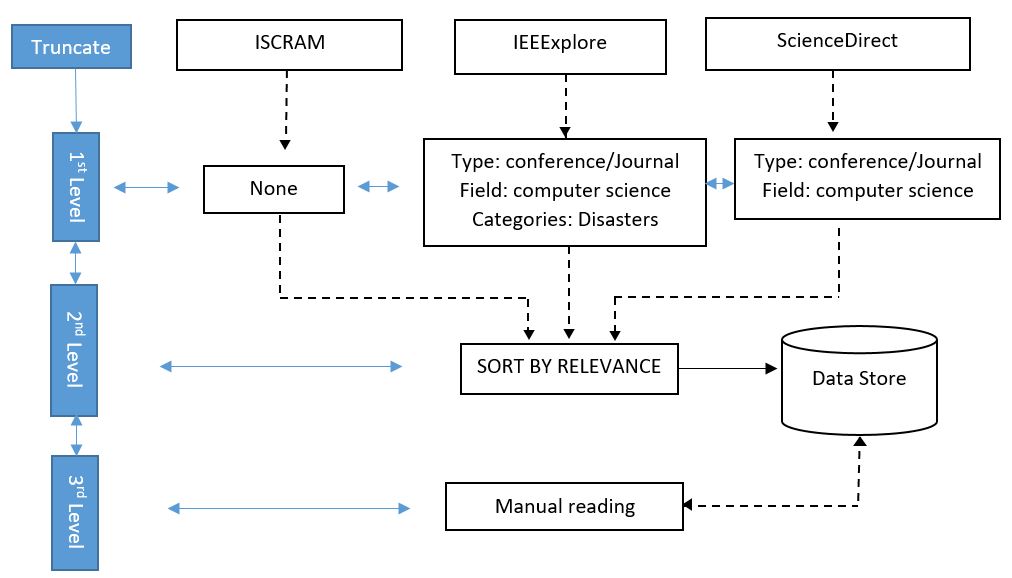}
  \caption{Inclusion and Exclusion flow diagram}
  \label{fig: truncate}
\end{figure}

The ISCRAM result did not require further reduction because we received only 76 articles. Reading through all of the resulting articles would be nearly impossible and time-consuming. Therefore, we attempted to utilize a systemic approach already available from the various servers to select the first 30 relevant articles rather than random sampling. The relevance reflects the articles that have more contributions to the body of knowledge and the frequency of citations. We ended up with 90 to balance the equation between the three databases. Further selection was carried out by means of manually reading the 90 articles as demonstrated in \autoref{fig: truncate} to ensure that each article made use of a machine learning method, had a crisis response matching, and appropriate research questions that addressed crises. To do this, we employed the criteria below:


\begin{enumerate}
  \item Include: based on the abstract that demonstrates a well-defined methodology
  \item Include: based on a conclusion that identifies at least one of the search terms as well as a metric that shows evidence for the study
  \item Exclude: based on a methodology that did not exemplify machine learning.
  \item Exclude: based on research that has no strong validation or premise for validation.
\end{enumerate}

\begin{table}[hbt!]
\begin{center}
    \caption{\label{tab:col-form}Data Item collection form.}
    \begin{tabular}{|p{1cm}|p{1.5cm}|p{4.5cm}|}
         \hline
         Index&Fields&Description\\
         \hline
         \hline
         DI-1&Title&The title of the article\\
         \hline
         \hline
         DI-2&Year&The publication year of the article\\
         \hline
         \hline
         DI-3&Database&The source of the publication \\
         \hline
         \hline
         DI-4&Techniques&The machine learning approach used in the article \\
         \hline
         \hline
         DI-5&Research Fields&The area of interest covered by the research(Computer Science) \\
         \hline
    \end{tabular}
\end{center}
\end{table}

\autoref{tab:col-form} illustrates the components of each manuscript that were extracted. This not only demonstrates the connection to the research questions but also provides a mechanism to confine data extraction to only the fields that are relevant. The title, year, database, machine learning techniques, and research questions covered are among such fields.

\subsection{Data Extraction}

The researchers manually coded the data in order to answer our two research questions. The labeling was done in four batches; methods, tasks, keywords, and algorithms, respectively. According to the selection criteria, all the publications considered for this study focused on crisis response and the application of machine learning technologies. 

\section{Result}

\begin{figure}[hbt!]
     \centering
     \begin{subfigure}[b]{0.22\textwidth}
         \centering
         \includegraphics[width=\textwidth]{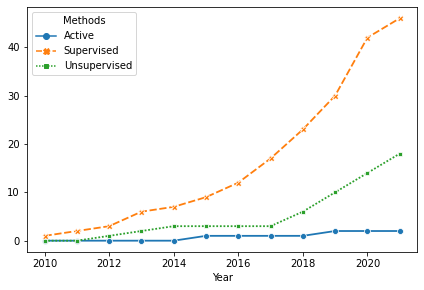}
         \caption{ML methods}
         \label{fig: methods}
     \end{subfigure}
     \begin{subfigure}[b]{0.21\textwidth}
         \centering
         \includegraphics[width=\textwidth]{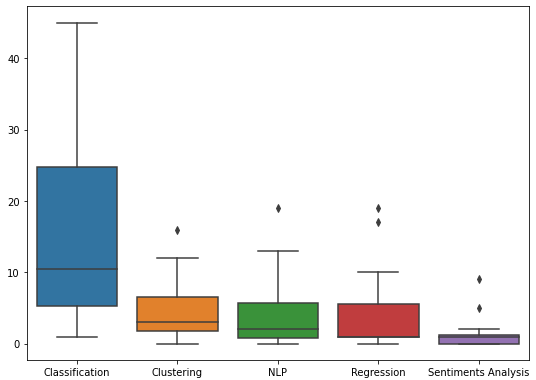}
         \caption{ML Tasks}
         \label{fig: task}
     \end{subfigure}
        \caption{Graph representation of the ML methods and tasks}
        \label{fig:mettask}
\end{figure}

From the analysis, majority of the articles reviewed supported the classification machine learning task more than the regression or clustering tasks. \autoref{fig: task} shows that 41\% of the reviewed papers made use of the classification task, while regression and clustering accounted for 18\% and 16\%, respectively. As shown in \autoref{fig: methods}, the supervised machine learning method garnered 69\% when mapped together with unsupervised and active learning, with 27\% and 3\%, respectively. This does not imply that one method is superior to the other, rather it demonstrates a preference  across scientific communities. The method preference of supervised learning can be attributed to the availability of training data peculiar to crisis from \textbf{CrisisNLP} \cite{alrashdi2019deep}, \url{https://crisisLex.org} - CrisisLexT26, SoSItalyT4, and BlackLivesMatterU/T1 \cite{olteanu2014crisislex}. Sometimes, the format in which crisis data is communicated may not be easily suited as a corpus for training a model. In this case, the unsupervised learning approach comes in handy \cite{arachie2020unsupervised}. Some of the studies examined had hybrid approaches where supervised and unsupervised methods were used depending on the availability of training data for some subset of their analysis. Active learning seems not to be largely applied in crisis research. One factor that could have influenced this is that active learning is a form of semi-supervised learning that engages outside sources to label a dataset. Because of that, the practice of semi-supervised learning was not noticed on a full scale. Following the distribution timeline of methods, it is clear that there is an upsurge in the use of supervised machine learning as the years go by across the three databases, as shown in \autoref{fig:mettask}.



The methods in \autoref{fig: methods} were further broken down into classification, clustering, regression, NLP, and sentiment analysis in \autoref{fig: task}. The NLP and sentiment analysis were identified separately in this figure because they can be a subclass of either classification, clustering, or regression tasks. It should be noted that sometimes they exist on their own (as transformers), not following the algorithmic process in the three classes mentioned above.

\begin{figure}
  \centering
  \includegraphics[width=.85\linewidth]{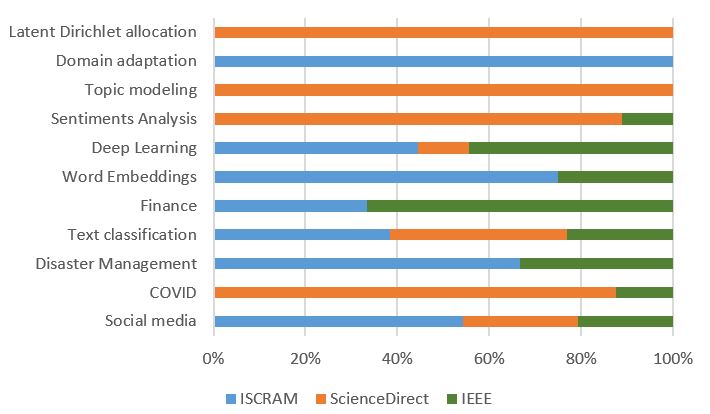}
  \caption{Distribution of keywords across reviewed articles}
  \label{fig: keywords}
\end{figure}

\begin{figure}
  \centering
        \includegraphics[width=1\linewidth]{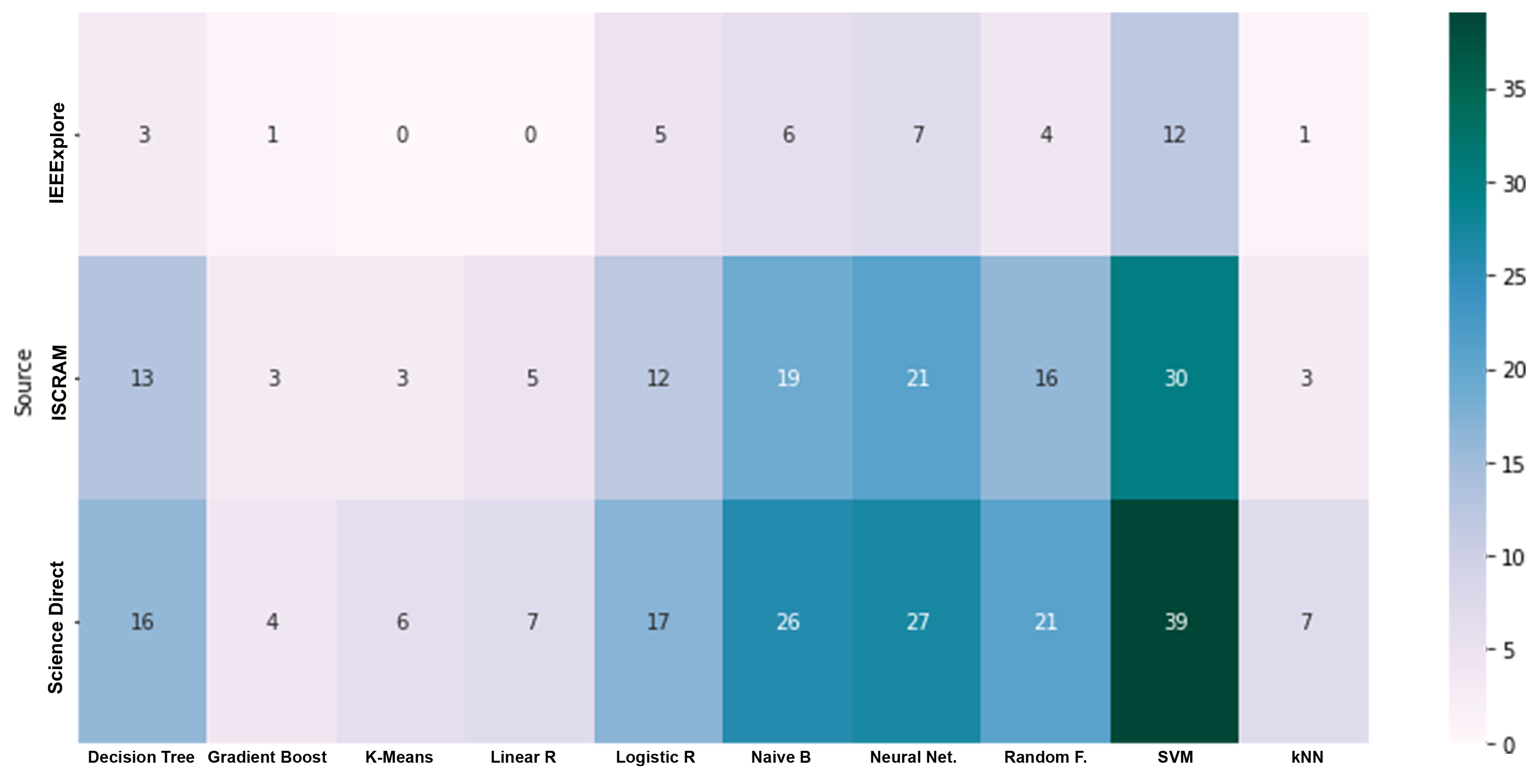}
  \caption{Algorithms applied in reviewed articles (heatmap)}
  \label{fig: algo1}
\end{figure}

\begin{figure}
  \centering
        \includegraphics[width=0.85\linewidth]{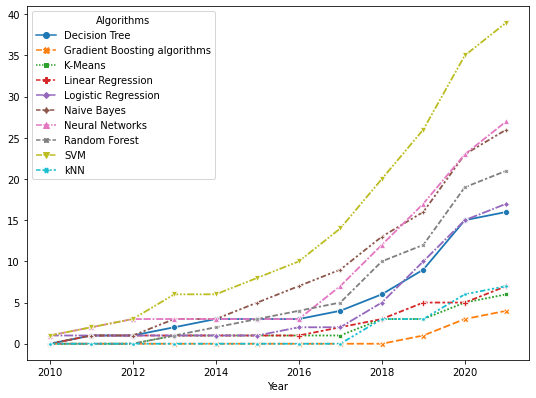}
  \caption{Machine learning algorithms applied in reviewed articles (Time series)}
  \label{fig: algo}
\end{figure}

The consistency in the usage of classification further strengthens the earlier statement about supervised learning being predominant amongst researchers. The next in line was clustering where k-nearest neighbors (kNN), K-Means, and some tasks with random forest and decision trees algorithms conform as shown in \autoref{fig: algo1} and \autoref{fig: algo}. The Regression tasks like linear and logistic regression shows that they are still relevant in today's research gaining 4\% and 10\% respectively. The analysis also shows that the implementation of machine learning methods for crisis management or evaluation reached its highest in the years 2020 and 2021. Again, the increase in the usage of machine learning classification techniques can be linked to the availability of training datasets, the ease with which the algorithms can be implemented, and the understandability of training data labels. The use of complex technologies can also be a contributing factor since such analysis may take human agents a considerable amount of time to analyze. The need for automated data processing, predictions, or analysis is something that has come to stay to aid humans.

The distribution of keywords in \autoref{fig: keywords} yielded some noteworthy results. The terms "COVID, disaster management, and social media" appeared more frequently in the reviewed publications, and they peaked in the year 2020. There was a clear pattern in the frequent mentions of social media and how they are used to aggregate crisis data. Our analysis also identified "social media" as the key term that received the most attention in the years 2019, 2020, and 2021, indicating that social media is a useful tool for gathering information about crises in the modern era. We can link the mention of a health crisis and COVID in 2021 to the recent COVID-19 pandemic and the way researchers are deploying intelligent systems to aggregate data across various media. Disaster management was also used as a keyword in several articles but was scarce in the year 2021. The mention of text classification, domain adaptation, and topic modeling was noted as well, and these relate to sub-methods in analyzing crises. The topic modeling which was scarcely present but important in the machine learning community was notably used in some research. This can be connected to our earlier statement that an NLP task can act as a transformer in an unsupervised environment (e.g., top modeling makes use of unsupervised methods).


Another significant discovery from this study is that when we further break down the methods to their algorithmic standards or the terminologies that indicate how they function, our earlier resolve could be strengthened. The SVM algorithm appears to dominate the reviewed articles, followed by the neural network. This also strengthens our earlier statement that the classification task and the supervised method were predominant. The strength of SVM can be uncovered in structural risk minimization, efficient memory management during training, and high dimensional spaces needed in datasets. Memory management is an issue for deep learning algorithms, which have more accuracy than SVM depending on the volume of data, but it seems the benefits of SVM appeal more to researchers in the crisis domain. The neural network was also prominent among the articles studied. Amongst the variations of neural networks present in crisis research are convolutional neural networks and long-term memory (LSTM) had more traction. The Naive Bayes, random forest, decision trees, and logistic regression also show promise as identified in \autoref{fig: algo}.



\section{Discussion}
We conducted this evaluative review on scientific articles published between 2010 and 2021, with an emphasis on crisis management and machine learning. Crises can emerge in different forms, e.g., health crises, natural disasters, economic crises, food crises, and political events, among others. These variations introduce complexity in developing automated methods to manage crises as well as track emerging events to allow for fast decision-making. The methodology was explicit enough in describing the meta-review processes and how we collected and evaluated different publications that addressed machine learning for crisis management. The concept of machine learning from the literature review and result sections is demonstrative of how many ML techniques and algorithms can be used to manage crises. It is evident from our results that all the algorithms had their fair representation in terms of the value they added. The machine learning field is continuously evolving as it tries to help in the analysis of large chunks of data, easing the tasks of data scientists in an automated process and changing the way data extraction and interpretation work.

All the reviewed articles produced results based on the structure of the problem they addressed, the type of crisis tackled, the source of the data, and the volume of the data. The percentage distribution, as shown in \autoref{fig: methods}, describes the preferred machine learning tasks. NLP tasks such as sentiment analysis, which may be classified as supervised or unsupervised learning, were highlighted as critical in automating crisis management. The classification, clustering, and regression tasks were highly preferred, with classification topping the list due to the availability of training datasets and easy implementation of the algorithms through pre-packaged libraries in popular programming languages like Python, R, and Java. Furthermore, there is a general recognition of a reproducibility crisis in science right now. Machine learning techniques are often simpler to perform analysis with. The reproducibility crisis is the increasing number of study findings that aren't replicated when a different group of researchers performs the same experiment. This problem has ramifications in a variety of sectors where machine learning is utilized to make discoveries.




\subsection{Limitations}

This study did not review literature from all the databases of scientific studies; instead, we culled articles from three databases that met the inclusion and exclusion criteria. As a result, our reviewed studies reflect research on public actions, crises, and machine learning from three venues. The results of this study may have been influenced by the search strategy employed in the paper, the researcher's biases, the unequal distribution of published journals or conference proceedings, and data extraction misrepresentation.

Both automated and manual search techniques were used in this study. Hundreds of data points were found as a result of the first iteration as seen in \autoref{fig: truncate}. The content of the research papers was used to inform the manual search procedure after the initial search. The possible studies were chosen and analyzed by three researchers. It is possible that relevant studies were skipped in the search results. As a result, the scope of this review may be constrained. Consequently, the validity of this study is limited to the 55 key papers included in this evaluative review.


\section{Conclusion}

Our findings show that a significant proportion of articles (41\%) used classification over regression or clustering, owing to the availability of training data/corpus and pre-packaged machine learning libraries. 69\% of the articles made use of the supervised machine learning method (RQ1), showing preference across scientific communities in dealing with crises. Consequently, 27\% of the studies made use of the unsupervised learning technique, while the remaining 4\% used active learning methods. To address RQ2, our analysis revealed the machine learning methods and prevalent keywords used in the reviewed articles. It suggests that the SVM, Neural Networks, Naive Bayes, and Random Forest algorithms, amongst others, are popular among researchers in the crisis management domain (RQ2). Also on RQ2, the keyword crisis informatics garnered great interest in the scientific literature explored. Some interesting projections like health and disaster management rose by the year 2020 and social media received the most attention in 2019, 2020, and 2021, implying that social media is beneficial to gathering crisis-related data in modern times.

\bibliographystyle{IEEEtran}
\bibliography{ieee}
\end{document}